\documentclass[sn-mathphys-num]{sn-jnl}

\usepackage{graphicx}%
\usepackage{multirow}%
\usepackage{amsmath,amssymb,amsfonts}%
\usepackage{amsthm}%
\usepackage{mathrsfs}%
\usepackage[title]{appendix}%
\usepackage{xcolor}%
\usepackage{textcomp}%
\usepackage{manyfoot}%
\usepackage{booktabs}%
\usepackage{algorithm}%
\usepackage{algorithmicx}%
\usepackage{algpseudocode}%
\usepackage{listings}%
\UseRawInputEncoding
\usepackage{caption,subcaption}
\usepackage[switch]{lineno}

\usepackage{colortbl}
\usepackage{xcolor}
\renewcommand{\textcolor}[2]{#2}
\theoremstyle{thmstyleone}%
\theoremstyle{thmstyletwo}%

\theoremstyle{thmstylethree}%

\begin{document}
\title[Article Title]{Mobility-Aware Federated Self-Supervised Learning in Vehicular Network}


\author[1]{\fnm{Xueying} \sur{Gu}}\email{xueyinggu@stu.jiangnan.edu.cn}

\author*[1,2]{\fnm{Qiong} \sur{Wu}}\email{qiongwu@jiangnan.edu.cn}

\author[3]{\fnm{Qiang} \sur{Fan}}\email{qf9898@gmail.com}

\author[4]{\fnm{Pingyi} \sur{Fan}}\email{fpy@tsinghua.edu.cn}

\affil[1]{\orgdiv{School of Internet of Things Engineering}, \orgname{Jiangnan University}, \orgaddress{\city{Wuxi}, \postcode{214122},  \country{China}}}

\affil[2]{\orgdiv{State Key Laboratory of Integrated Services Networks}, \orgname{Xidian University}, \orgaddress{\city{Xi'an }, \postcode{710071}, \country{China}}}

\affil[3]{\orgdiv{Qualcomm}, \orgname{San Jose CA}, \orgaddress{\postcode{95110}, \country{USA}}}

\affil[4]{\orgdiv{Department of Electronic Engineering, Beijing
		National Research Center for Information Science and Technology}, \orgname{Tsinghua
		University}, \orgaddress{\city{Beijing}, \postcode{100084}, \country{China}}}


\abstract{\textcolor{red}{The development of the Internet of Things (IoT) has led to a significant increase in the number of devices, generating vast amounts of data and resulting in an influx of unlabeled data. Collecting this data enables the training of robust models, supporting a broader range of applications. However, labeling these data can be costly, and models dependent on labeled data are often unsuitable for rapidly evolving fields like vehicular networks or mobile Internet of Things (MIoT), where new data continuously emerges. To address this challenge, Self-Supervised Learning (SSL) offers a way to train models without the need for labels. Nevertheless, the data stored locally in vehicles is considered private, and vehicles are reluctant to share it with others. Federated Learning (FL) is an advanced distributed machine learning approach that protects each vehicle's privacy by allowing models to be trained locally and exchange the model parameters across multiple devices simultaneously.} Additionally, vehicles capture images while driving through cameras mounted on their rooftops. If the vehicle's velocity is too high, images, donated as local data, may become blurred. Simple aggregation of such data can negatively impact the accuracy of the aggregated model and slow down the convergence speed of FL. This paper proposes a FL algorithm based on image blur levels for aggregation, called FLSimCo. This algorithm does not require labels and serves as a pre-training stage for SSL in vehicular networks. Simulation results demonstrate that the proposed algorithm achieves fast and stable convergence.}




\maketitle

\section{Introduction}\label{sec1}

The development of Internet of Things (IoT) makes many practical applications available to people, such as automatic navigation, weather forecast and self-driving systems, which improves the happiness of life \cite{wsy01,ss01,ldx01}. Training a good model can make the practical application more robust, which requires a lot of data. Moving vehicles can constantly collect new data by their devices, most of which can be captured in the form of images via camera mounted on the roof of vehicles. After the acquisition is completed, the vehicle process the image data to realize image classification. It provides the necessary information for self-driving and driver assistance systems, and helps drivers perceive and understand their surroundings \cite{canagnostopoulos2023,wwh,wwh1}. Therefore, image classification in self-driving plays a key role in reducing the risk of traffic accidents and improving road safety.

As mentioned earlier, training a robust model requires a lot of data. However, the data stored locally on each vehicle may be private and sensitive, and drivers are reluctant to share it with others.
\textcolor{red}{Due to the single driving environment, the categories of images captured by each vehicle may be shewed toward one or two classes. Using these images for local training may result in local models bias and lack of integrity \cite{wxb01,xsy01}. }Compared to single vehicle training, FL can solve this issue by aggregating different vehicles' local trained models. This approach can take a large data set into account in the training process without the need for vehicles to share local data, resulting in superior performance and enhanced generalization \cite{ryan2023,szy1, szy2}. 

\textcolor{red}{However, employing FL in classification still faces the following two issues:} low image quality and missing or incorrect labels. The first one is due to the motion blur caused by the movement of vehicles, and the other is because of the high cost and incorrectness of labeling. In a vehicular network, a high velocity usually leads to insufficient exposure time for the camera sensor, causing motion blur. In this paper, we consider the motion blur caused by different vehicle velocities in the training process, and adjust corresponding weights of the model parameters for FL model aggregation. \textcolor{red}{Move over, Self-supervised learning (SSL) can abandon labels for pre-training, and thus mitigates the impact of incorrect labels on the model and removes the cost of the labeling process, which makes it suitable for mobile IoT (MIoT)\cite{czhang2022dualtemp}}.

To the best of our knowledge, few research works have taken into account both the privacy protection of vehicles and the blurring of images in real scenes, as well as the cost of labels during the training of models.

The remaining of this paper is organized as follows. In Section \ref{sec2}, we will review related works. The Section \ref{sec3} details the system model we designed. In Section \ref{sec4}, we detail the process of our proposed FLSimCo. Section \ref{sec5} will showcase and discuss the experimental results obtained in the simulation environment. Finally, we will summarize the research findings and present conclusions in the Section \ref{sec6}.

\section{Review of related works}\label{sec2}

In recent years, the emergence of deep neural networks, particularly convolutional neural network (CNN), has facilitated significant advances in computer vision benchmarks. SSL is a special form of unsupervised learning, and its main feature is that it uses information inherent in the data itself to help machine learning models get a better representation, thereby improving performance across a variety of tasks.

Until now, research on unsupervised learning has focused on mining shared features between pre-trained tasks and downstream tasks \cite{cdoersch2016}. In  \cite{zwu}, Wu \emph{et al.} introduced noise contrast estimation (NCE) loss as an objective function to distinguish between different instances. Each image was treated as a positive sample and the others as negative samples, effectively treating each image as a category, constituting an instance-level classification task. In \cite {tchen2020SimCLR, skong2023}, Chen \emph{}et al. proposed SimCLR, which maximized the sample characteristics of similarity to study representations. It abandoned the memory library and generated more negative samples by increasing the batch size \cite{rhjelm}. The core idea of these methods is to encourage models to place similar data representations (positive sample pairs) together in the embedding space, while separating dissimilar representations (negative sample pairs) to learn feature extraction.
Since these methods do not consider the fact that the distance between samples of the same category should be smaller than that of different categories' samples, but treat all pairs equally. In this way, a large number of negative samples are required. It will pose significant challenges to storage and computing power in the vehicle environment, increasing the requirements of hardware configuration levels.

MoCo, as a self-supervised learning method for CNN, has made an important contribution by introducing queue and momentum updating techniques to create a large and consistent dictionary conducive to contrast learning \cite{khe2020,xchen2020MoCoV2,jzhao01}. It replaced the original memory library with a queue as an additional data structure to store these negative samples. Momentum encoder $k$ is used to replace the original loss constraint term. The key advantage of the MoCo family is the use of the queue as negative samples and the momentum update of the queue, which greatly reduces the need for vehicle storage capacity and computing power in local training. In \cite{tcai2022}, Cai \emph{et al.} used MoCo for the pre-training model weights on ImageNet and achieves excellent results in downstream fine-tuning tasks. In \cite{jzhao2021Hotfed}, Zhao \emph{et al.} employed MoCo for unsupervised learning with large amounts of unlabeled data on remote servers. The local vehicles downloaded the trained model and used it as the initialization model.
\begin{figure}
	\center
	\includegraphics[scale=0.35, trim=2cm 1cm 2cm 2cm, clip]{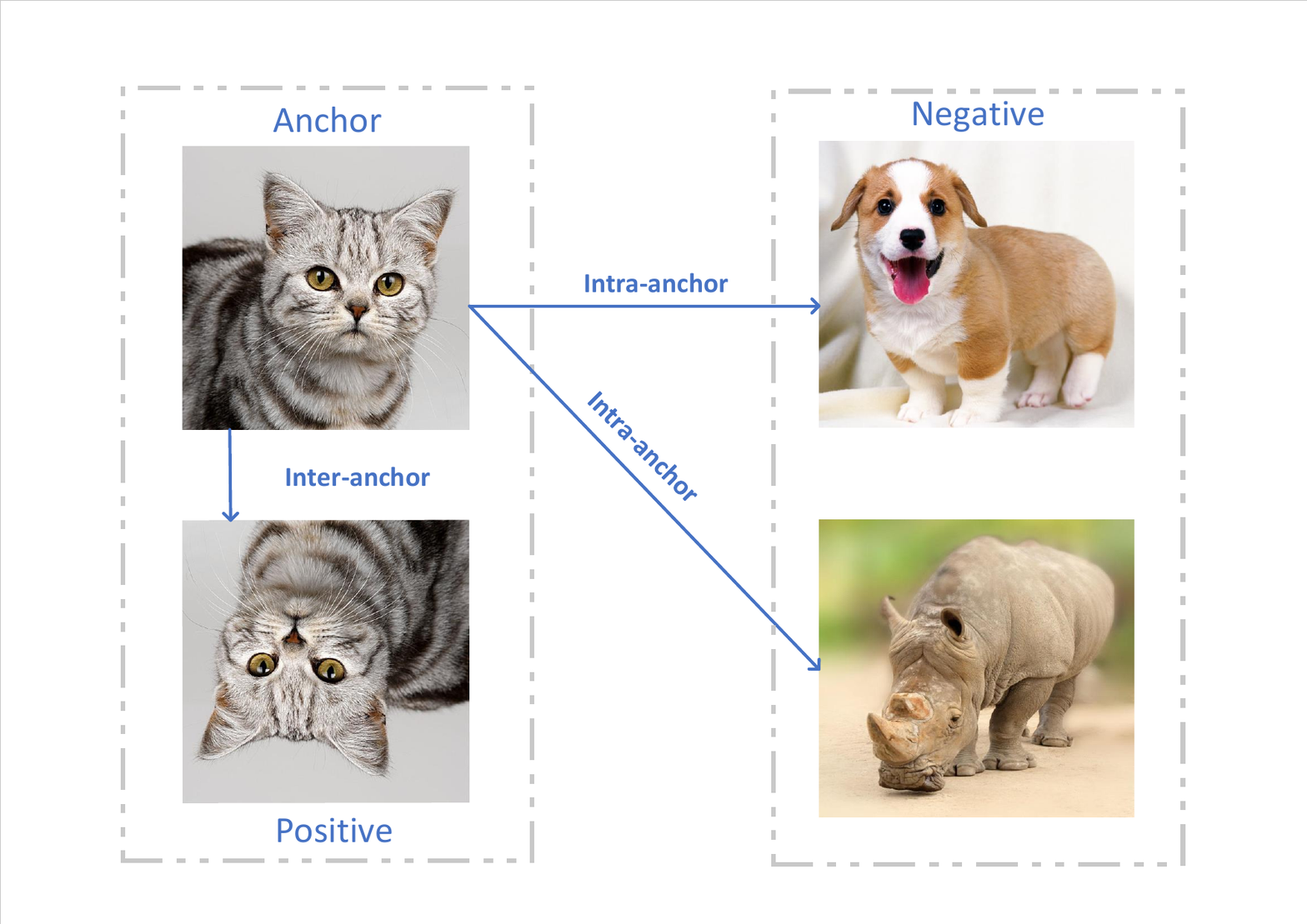}
	\caption{Inter-anchor sample and intra-anchor samples}
	\label{fig1}
\end{figure}

These methods are characterized by the presence of a large default datasets that can be used for unsupervised learning, and the accuracy increases with the expansion of the dictionary size. However, it is important to note that, larger dictionaries also requires larger memory size, increasing the storage cost and the computing capability of hardness. In \cite{czhang2022dualtemp}, Zhang \emph{et al.} proposed SimCo, which attributed the need for a large dictionary in MoCo to hardness awareness between anchors, and believed that the consistency between positive and negative keys was more crucial than that between negative keys. Therefore, it utilized dual temperatures to differentiate inter-anchor samples and intra-anchor samples, as shown in Fig. \ref{fig1}, which eliminates the need for a dictionary of negative samples.

Due to the imbalance of data categories, the data distribution faces the challenge of Non-Independent Identically Distributed (Non-IID) characteristics, requiring more training rounds and interaction frequency to improve accuracy \cite{jzhao00}. However, increased interaction means require more computing and network resources \cite{jzhao02,zhangwenjun}. To protect the privacy of local data and its distribution, FedCo proposed to employ FL to aggregate local models. Each vehicle passed its own $k$ value to the RSU for forming a bigger new queue. However, a new queue consisting of the $k$ values of different vehicles violates MoCo's inherent requirement for consistency in negative key pairs. When the vehicle uploaded the trained model and the corresponding $k$ value, it can, to some extent, reconstruct the original input. As such, it does not ensure privacy protection and defeats the original purpose of using FL. In\cite{mfeng2022}, Feng \emph{et al.} used federated SSL for a single-event classification, adding a binary classifier for each new event by adhering to a one-to-many paradigm. However, these FL algorithms do not take into account the effect of image blur. As a result, they cannot effectively simulate real-world scenarios.

In this paper, we will train the local model on the vehicles side by SSL with dual temperatures. After the training is completed, only the local models of vehicles are uploaded, and then Road Side Unit (RSU) will aggregates the received local model, which not only protects the privacy of vehicles, but also produces a model with better transform through distributed training. At the same time, considering that the images collected in the actual vehicular network may be blur, the blur level is used as the weight during aggregation, which makes the aggregated model more reliable and stable.

\begin{figure}
	\centering
	\includegraphics[scale=0.35, trim=1cm 1cm 1cm 0.5cm, clip]{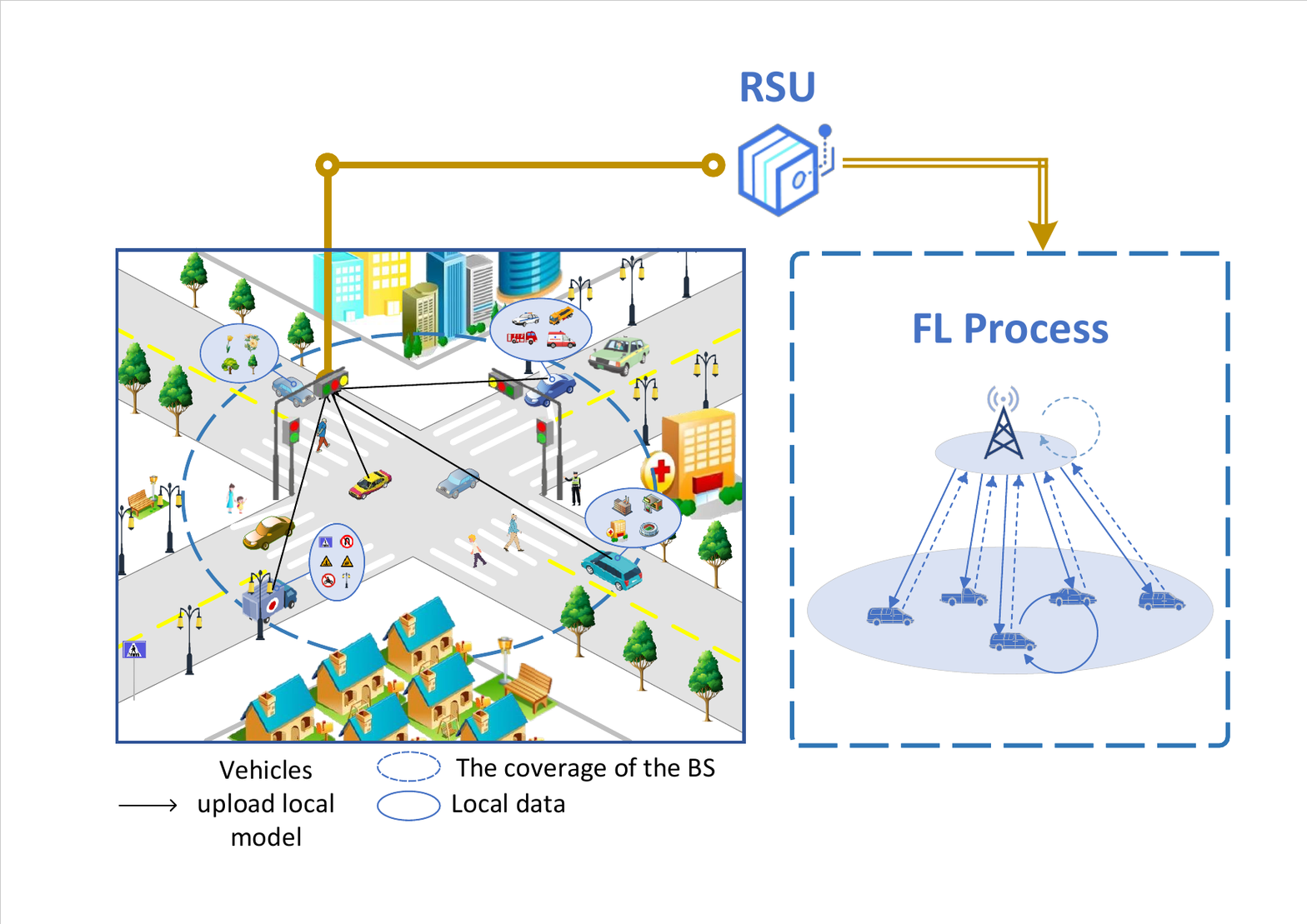}
	\caption{System scenario }
	\label{fig2}
\end{figure}
\section{System model}\label{sec3}

\subsection{System design}\label{subsec2}

As shown in Fig. \ref{fig2}, we consider a scenario where an RSU is deployed at the intersection and there are several vehicles driving in coverage of RSU. Each vehicle driving straightly when cross the intersection. We firstly build a truncated Gaussian distribution model as the mobility model for each vehicle. At the same time, each vehicle is equipped with a large mount of images, donated as local data, captured by camera mounted in the roof before entering the coverage of the RSU. It is worth noting that vehicles coming from different directions may capture different categories of images from each other.

In each round of FL,  each vehicle downloads the parameters of global model from the RSU as it enters the coverage of the RSU. Then each vehicle sets the parameters of global model to local model for the pre-training of SSL to classify each category of captured local data. During this process, the vehicle randomly selects a specified number of images from its local data. When the local training is complete, each vehicle uploads the parameters of local model to RSU. Then, the RSU  aggregates the local models from different vehicles to obtain a new global model. After that, FL moves on to the next round until the maximum round $R^{max}$ is reached. Next, we will describe the models for each round in the system.

\subsection{Mobility model}
Similar with \cite{qwu2023}, we adopt the following mobility model to reflect the real vehicle mobility. We consider the velocities of different vehicles are Independent Identically Distributed (IID). The velocity of each vehicle follows the truncated Gaussian distribution. Let $N_r$ be the number of vehicles traveling within the coverage area of RSU in the round $r$, $v_{n_r}$ be the velocity of vehicle $n_r\in[1,\ N_r]$, and $v_{min}$ and $v_{max}$ be the minimum and maximum velocity of vehicles, respectively, namely velocity $v_{n_r}\in\left[v_{min},\ \ v_{max}\right]$, and the probability density function of $v_{n_r}$ is expressed as  \cite{zyu2021}
\begin{equation}
	\vspace{-0.25cm}
	f\left(v_{n_r}\right) = \left\{ \begin{aligned}
		\frac{{{e^{ - \frac{1}{{2{\sigma ^2}}}{{({v_{n_r}} - \mu )}^2}}}}}{{\sqrt {2\pi {\sigma ^2}} [erf(\frac{{{v_{\max }} - \mu }}{{\sigma \sqrt 2 }}) - erf(\frac{{{v_{\min }} - \mu }}{{\sigma \sqrt 2 }})]}}\\
		{v_{min}} \le {v_{n_r}} \le {v_{max}},\\
		0 \qquad \qquad \qquad \qquad \quad otherwise.
	\end{aligned} \right.
	\label{eq1}
\end{equation}where $erf(\mu, \sigma^2)$ is the Gaussian error function of velocity $v_{n_r}$ with mean $\mu$ and variance $\sigma^2$.

\section{FLSimCo algorithm}
\label{sec4}

We establish a novel algorithm called FLSimCo, which means simplified MoCo with no momentum in FL. \textcolor{red}{Before the whole training start, RSU stores a global model. Each vehicle stores a encoder and M images, donated as local model and local data. }Typically, the specific FLSimCo process is as follows:

\textbf{\emph{Step 1, initialization}}: The RSU randomly initializes the parameters of global model $\theta^0$.

\textbf{\emph{Step 2, local training}}: \textcolor{red}{For round $r$, $N_r$ vehicles take part in the FL, where each vehicle $n_r$ downloads the model $\theta^r$ from RSU and set parameter of $\theta^r$ to local model $\theta_{n_r}$. }

The blur level $L_{n_r}$ of local image data in the vehicle $n_r$ can be represented as \cite{sshirmohammadi,jacortes-Osorio2018}:
\begin{equation}
	L_{n_r}=\frac{Hs}{Q}v_{n_r},
	\label{eq4}
\end{equation}where $\frac{Hs}{Q}$ is the parameter of the camera, in which $H$ represents the exposure time interval, $s$ is the focal length, and $Q$ represents pixel units.

For each image $x_r^i\in\left\{x_r^1,\ x_r^2,\ \ldots x_r^i,\ldots x_r^M\right\}$, applying two different data augmentation methods $\pi_1(\cdot)$ and $\pi_2(\cdot)$. \textcolor{red}{$\pi_1(\cdot)$ performs a horizontal flip on the image with a 50\% probability, followed by converting the image to grayscale with a 20\% probability. $\pi_2(\cdot)$ randomly alters the image's brightness, contrast, saturation, and hue with an 80\% probability (each within a range of 0.4), followed by converting the image to grayscale with a 40\% probability. It is also noted that $\pi_1\left(\cdot\right)$ and $\pi_2\left(\cdot\right)$ share the same original image when they process the image in different way. After that, the augmented images are passed through the encoder $f^r$ with the ResNet structure, which is the local model with parameters $\theta_{n_r}$. }Then we obtain anchor sample $q_{n_r}^i$, positive sample $k_{n_r}^i$ and negative samples $k_{n_r}^j$:
\textcolor{red}{
\begin{equation}
	q_{n_r}^i=f^r\left[\pi_1\left(x_r^i\right)\right],\ i\in\left[1,M\right],
	\label{eq5}
\end{equation}
\begin{equation}
	k_{n_r}^i=f^r\left[\pi_2\left(x_r^i\right)\right],\ i\in\left[1,M\right],
	\label{eq6}
\end{equation}
\begin{equation}
	k_{n_r}^j=f^r\left(x_r^j\right),\ j\in\left[1,M\right]\ and\ j\neq i.
	\label{eq80}
\end{equation}
}
According to \cite{czhang2022dualtemp}, the dual-temperature (DT) loss $\mathcal{L}_{q_{n_r}^i}^{DT}$ of the $i$-th image of vehicle $n_r$ anchor sample in the round $r$ can be calculated as

\begin{equation}
	\begin{split}
		\mathcal{L}_{q_{n_r}^i}^{DT}=-sg\left[\frac{{{(W}_\beta)}_{n_r}^i}{{{(W}_\alpha)}_{n_r}^i}\right]\times log\frac{\exp{\left(\frac{q_{n_r}^i\cdot k_{n_r}^i}{\tau_\alpha}\right)}}{\exp{\left(\frac{q_{n_r}^i\cdot k_{n_r}^i}{\tau_\alpha}\right)}+\sum_{j=1}^{K}\exp{\left(\frac{q_{n_r}^i\cdot k_{n_r}^j}{\tau_\alpha}\right)}},
	\end{split}
	\label{eq7}
\end{equation}
and
\begin{equation}
	{{(W}_\beta)}_{n_r}^i=1-\frac{\exp{\left(\frac{q_{n_r}^i\cdot k_{n_r}^i}{\tau_\beta}\right)}}{\exp{\left(\frac{q_{n_r}^i\cdot k_{n_r}^i}{\tau_\beta}\right)}+\sum_{j=1}^{K}\exp{\left(\frac{q_{n_r}^i\cdot k_{n_r}^j}{\tau_\beta}\right)}},
	\label{eq8}
\end{equation}
\begin{equation}
	{{(W}_\alpha)}_{n_r}^i=1-\frac{\exp{\left(\frac{q_{n_r}^i\cdot k_{n_r}^i}{\tau_\alpha}\right)}}{\exp{\left(\frac{q_{n_r}^i\cdot k_{n_r}^i}{\tau_\alpha}\right)}+\sum_{j=1}^{K}\exp{\left(\frac{q_{n_r}^i\cdot k_{n_r}^j}{\tau_\alpha}\right)}},
	\label{eq9}
\end{equation}where $K$ represents the queue length and $sg[\cdot]$ indicates the stop gradient. The denominator in the above equations consists of one positive sample and $K$ negative samples. It is noting that '$\cdot$' in Eq. \eqref{eq7} - Eq.  \eqref{eq9} means dot product. $\tau_\alpha$ and $\tau_\beta$ are different temperature hyper-parameters \cite{khou2020}, and controls the shape of the samples distribution.  Based on the different requirements for dictionary size by inter-anchor and intra-anchor, and the temperature's ability to control the feature distribution, different temperatures will be used to control the distance between different samples, thus eliminating MoCo's dependency on a large dictionary, which will remove the inter-anchor's reliance on a large dictionary. 

The objective function can be defined as minimizing the loss function. \textcolor{red}{ The final ideal value can be donated as ${\hat{\theta}}_{n_r}$, and ${\hat{\theta}}_{n_r}$ can be expressed as}
\begin{align}
	{\hat{\theta}}_{n_r}=\underset{\theta_{n_r}}{{\rm argmin}}\frac{1}{M}\sum_{i=1}^{M}{\mathcal{L}_{q_{n_r}^i}^{DT}\left(\theta_{n_r}, \ q_{n_r}^i,\ k_{n_r}^i,\ k_{n_r}^j\right)},
	\label{eq11}
\end{align}where $\theta_{n_r}$ represents the parameter of local model of vehicle $n_r$ in round $r$. \textcolor{red}{ Each vehicle performs the local training to approach ${\hat{\theta}}_{n_r}$ according to Stochastic Gradient Descent (SGD) algorithm, and the process in round $r$ can be expressed as}

\begin{equation}
	\theta_{n_r}\gets\ \theta_{n_r}-\eta^r\nabla\mathcal{L}^{DT}\left(\theta_{n_r},\ q_{n_r}^i,\ k_{n_r}^i,\  k_{n_r}^j\right),
	\label{eq12}
\end{equation}where $\eta^r$ represents the learning rate for the  round $r$.

It is worth noting that during the process of local training, vehicles also capture new images $\left\{x_{r+1}^1,\ x_{r+1}^2,\ \ldots x_{r+1}^i,\ldots x_{r+1}^M\right\}$, which will be designated as local data for round $r+1$.

\textbf{\emph{Step 3, Upload model}}: $N_r$ vehicles upload the parameter $\left\{\theta_1,\ \theta_2\ldots\theta_{n_r}\ldots\theta_{N_r}\right\}$ of local models after their local training finished, along with the velocity $\left\{v_1,v_2,\ldots v_{n_r}\ldots v_{N_r}\right\}$. Specifically, vehicle $n_r$ uploads trained parameters of local model $\theta_{n_r}$, along with the vehicle $v_{n_r}$ to RSU when local training is finished.

\textbf{\emph{Step 4, Aggregation and Update}}: After receiving the trained models from $N_r$ vehicles, the RSU employs a weighted federated algorithm to aggregate the parameters of $N_r$ models based on the blur level $L_{n_r}$. The expression for the aggregated model is
\textcolor{red}{
\begin{equation}
	\theta^{r+1}=\sum_{n_r=1}^{N_r}\left[\frac{\left(\sum_{n_r=1}^{N_r}L_{n_r}-L_{n_r}\right){\theta}_{n_r}}{\sum_{n_r=1}^{N_r}L_{n_r}}\right],
	\label{eq13}
\end{equation}where $\theta^{r+1}$ represents the new global model for round $r+1$.}

Repeat the above step 2 to step 4 until reaching max round $R^{max}$.

\section{Results}\label{sec5}

In this section, we will introduce the setup of experiments, show the results, and give a brief explanation.

\subsection{Experimental setup}

	\textcolor{red}{Python 3.10 is utilized to conduct the simulations, which are based on the scenarios outlined in the Section \ref{sec3}. }We adopt an improved ResNet-18 with a fixed dimension of 128-D as the backbone model and employ SGD as the optimizer. In addition, inspired by the concept of cosine annealing, we gradually reduce the learning rate at different stages of training to improve the training efficiency of the model. \textcolor{red}{Other simulation parameters are detailed in Table \ref{tab1}.}

	\begin{table}[htbp]
	\centering
	\caption{Hyper-parameter}
	\label{tab1}
	\arrayrulecolor{red}
	\begin{tabular}{|c|c|c|c|}
		\hline
		\textbf{Parameter} & \textbf{Value} & \textbf{Parameter} & \textbf{Value} \\
		\hline						  
		$\tau_\alpha$ & 0.1 & $\tau_\beta$ & 1 \\
		$\mu$ & 0.5 & $\sigma$ & 8\\
		$v_{\text{min}}$ & 16.67 m/s & $v_{\text{max}}$ & 41.67 m/s\\
		$\text{Total number of vehicles}$ & 95 & M & 520 \\
		$\text{Momentum of SGD}$ & 0.9 & $\text{Original learning rate}$ & 0.9\\
		$\text{Weight decay}$ & $5\times{10}^{-4}$ & $\text{MoCo momentum of updating key encoder}$ & 0.99\\
		$R^{max}$ & 150&&\\

		\hline
	\end{tabular}
\end{table}
\textcolor{red}{
\textbf{\emph{Testing}}: We rank the predicted labels based on their probabilities from highest to lowest. If the most probable predicted label (i.e., the top label) matches the true label, the prediction is considered correct, and this is referred to as the Top1 accuracy. Each experiment is conducted for three times, and the final result is the average of these three experiments.}

In the vehicle scene, it is easy for the vehicles to collect enough image data during driving. However, due to the limitations of the environment, storage and camera perspective, the categories of image data of each vehicle are limited, which may not meet the IID requirement \cite{khsieh2020,yzhao2022}. To be specific, when the vehicle uses the image data stored by itself for local training, a model shewed to its own image categories is obtained, namely a model with poor generation. Therefore, we will conduct our simulation on IID and Non-IID datasets.

\textbf{\emph{Datasets}}: \textcolor{red}{In intersection-related scenes, the categories of objects are relatively limited, and images of the same category of object are frequently collected. Therefore, CIFAR-10 is selected in order to check the working status of the proposed algorithm quickly.} The datasets CIFAR-10 with 50,000 unlabeled images as the training datasets, which is distributed in 10 different categories, i.e., each category consists of 5,000 images \cite{CIFAR-10}.
\begin{figure*}
	\centering
	\begin{subfigure}{0.32\linewidth}
		\includegraphics[width=\linewidth]{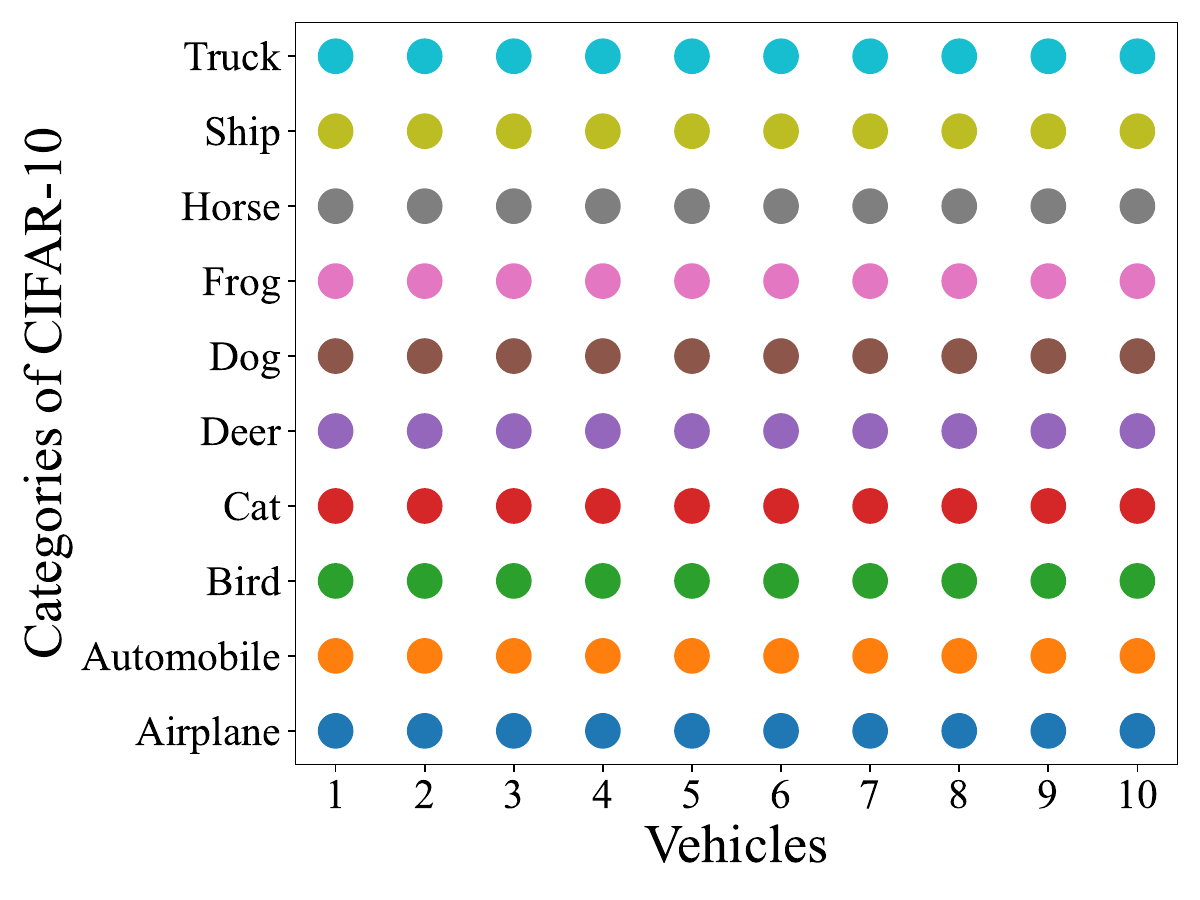}
		\caption{IID data}
		\label{fig4a}
	\end{subfigure}
	\hfill
	\begin{subfigure}{0.32\linewidth}
		\includegraphics[width=\linewidth]{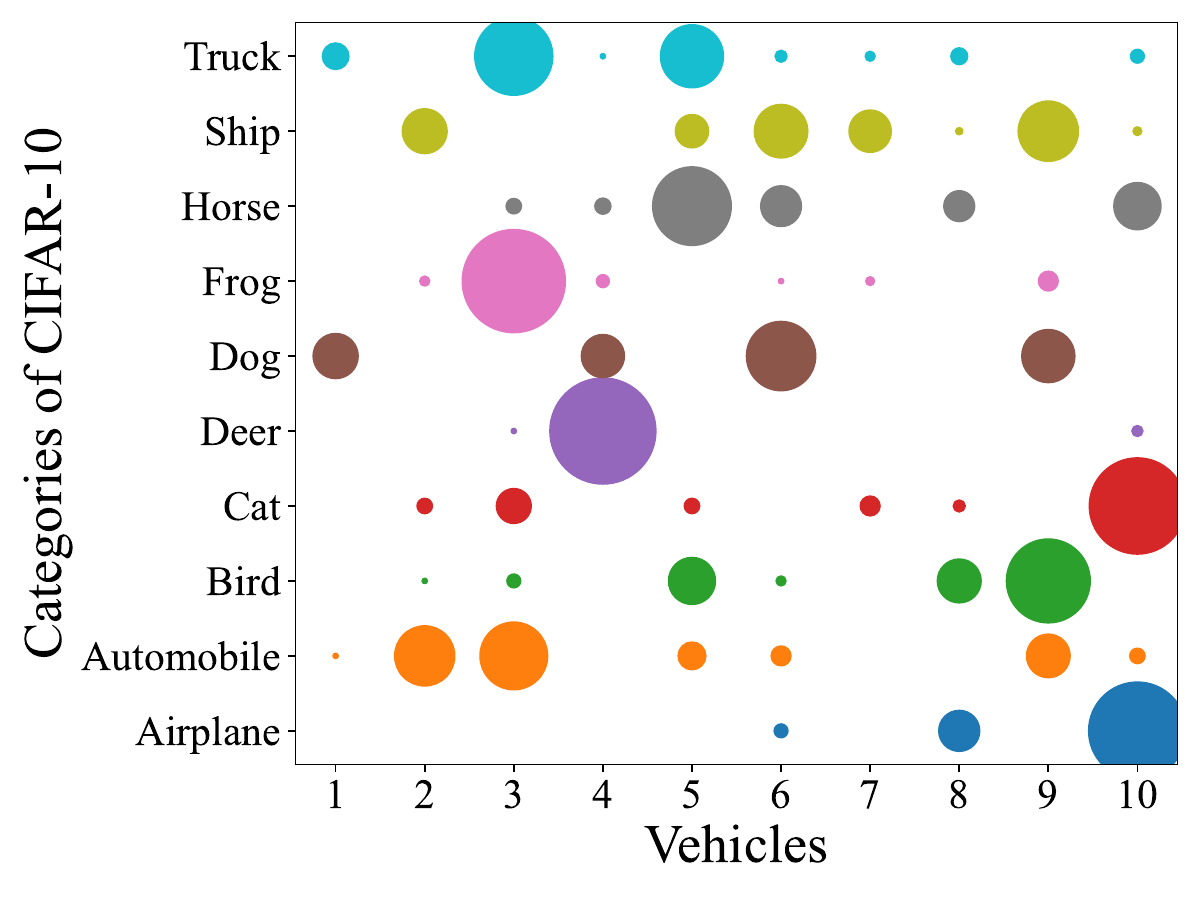}
		\caption{Non-IID with $\alpha$ = 0.1}
		\label{fig4b}
	\end{subfigure}
	\hfill
	\begin{subfigure}{0.32\linewidth}
		\includegraphics[width=\linewidth]{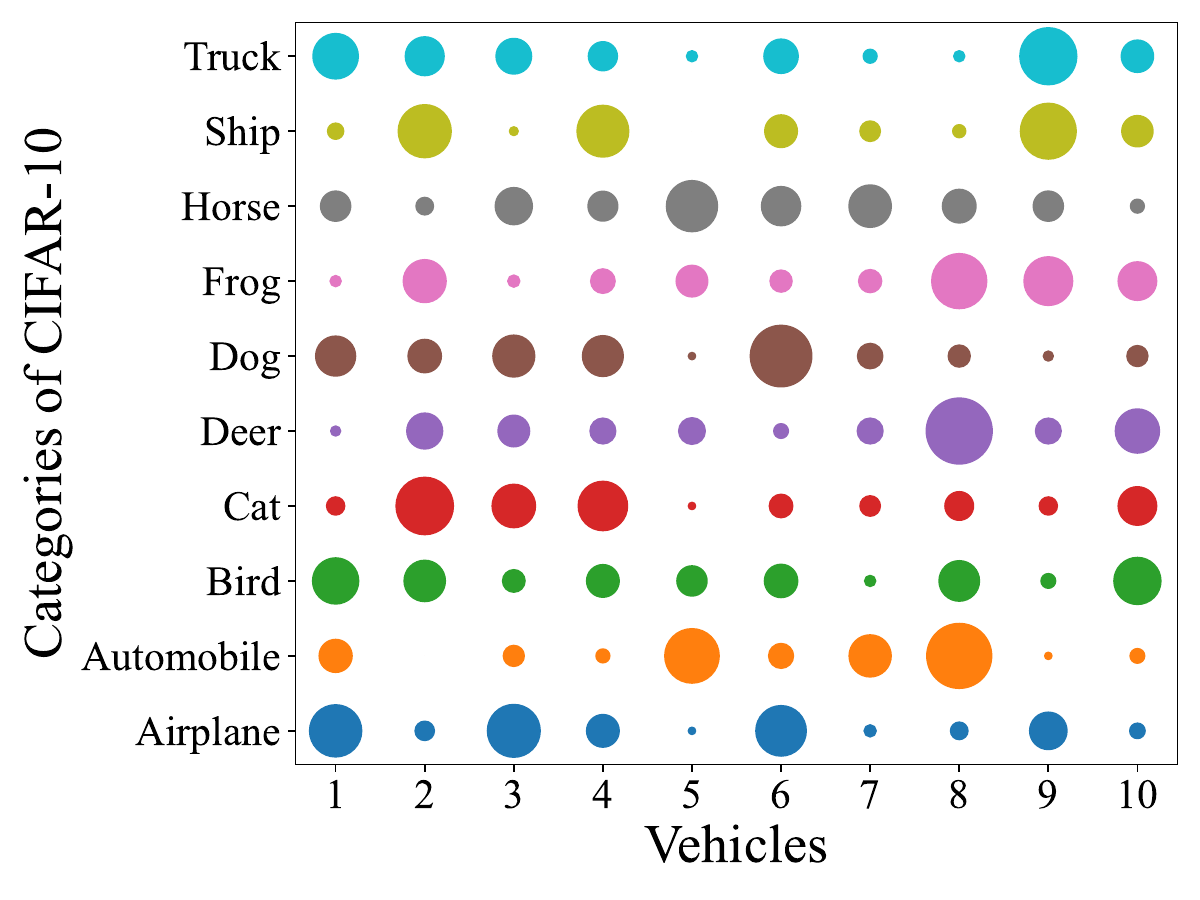}
		\caption{Non-IID with $\alpha$ = 1}
		\label{fig4c}
	\end{subfigure}
	\caption{Data Category Distribution Plot}
	\label{fig4}
\end{figure*}

(1) \textbf{\emph{IID}}: Samples that follow IID are independent of each other and share the same distribution. The appearance of each sample is not affected by the other samples. We uniformly assigned 10 categories from 50,000 images in CIFAR-10 to 95 vehicles, ensuring that each vehicle have at least 520 images available for training. As can be seen from the Fig. \ref{fig4a}, the categories on each vehicle are evenly distributed.

(2) \textbf{\emph{Non-IID}}: IID provides theoretical convenience, but in a practical scenario, there is very little data that meets the IID requirement. Therefore, it is importance to use data under Non-IID conditions to transfer the model to the real scenario. As shown in Fig. \ref{fig4b} and \ref{fig4c}, the Dirichlet distribution parameter $\alpha$ is 0.1 and 1, respectively \cite{zzhang2023}. Clearly, the smaller the $\alpha$, the larger the gap between data categories. We set the Dirichlet distribution parameter $\alpha$ to 0.1 to simulate the Non-IID data in the vehicular scene, in order to simulate the uneven distribution of the  image categories collected by each vehicle due to the limited viewing perspective and environmental constraints. In order to ensure that there is enough data for local training, for CIFAR-10 we ensure that there are at least 520 images per vehicle.

\subsection{\textcolor{red}{Simulation evaluation}}

According to FedCo \cite{sweiFedCo}, in the round $r$ of the training process, we set each vehicle uploads all stored $k$-values (with a batch size set to 512 in the experiment) to the RSU (global queue set to 4096) to update the global queue. 

\begin{figure}
	\center
	\includegraphics[scale=0.5]{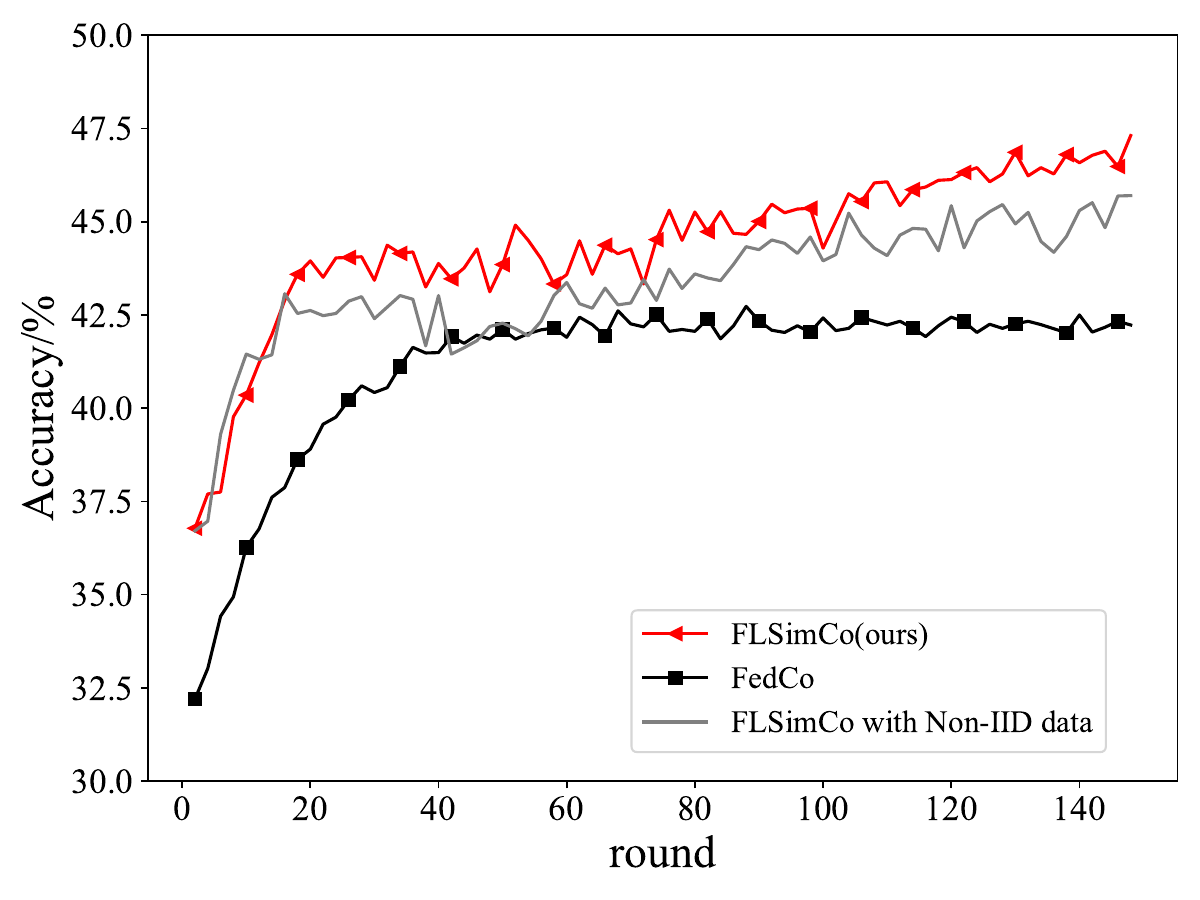}
	\caption{FLSimCo VS other methods}
	\label{fig5}
\end{figure}

As shown in Fig. \ref{fig5}, we compare our proposed FLSimCo with FedCo algorithm. Our proposed method FLSimCo is represented by a red line in the same diagram and outperforms FedCo given the same number of rounds. From FedCo's perspective, the update queues with $k$ values from different vehicles has compromised the Negative-Negative consistency requirement in MoCo, resulting in a less accurate approach. Meanwhile, FedCo enables the vehicle's own $k$ value to be uploaded to RSU, which also goes against FL's purpose of protecting user privacy. In addition, we also conducted experiments on Non-IID datasets, and the results show that the training performance of Non-IID data sets is slightly lower than that of IID data sets, but still better than FedCo algorithm. \textcolor{red}{Numerically, the FLSimCo method improves classification accuracy over the existing FedCo method by 13.03\% on IID datasets and by 8.2\% on Non-IID datasets. }

\begin{figure*}[!t]
	\centering
	\subfloat[The accuracy of Top1]{%
		\includegraphics[width=0.48\linewidth]{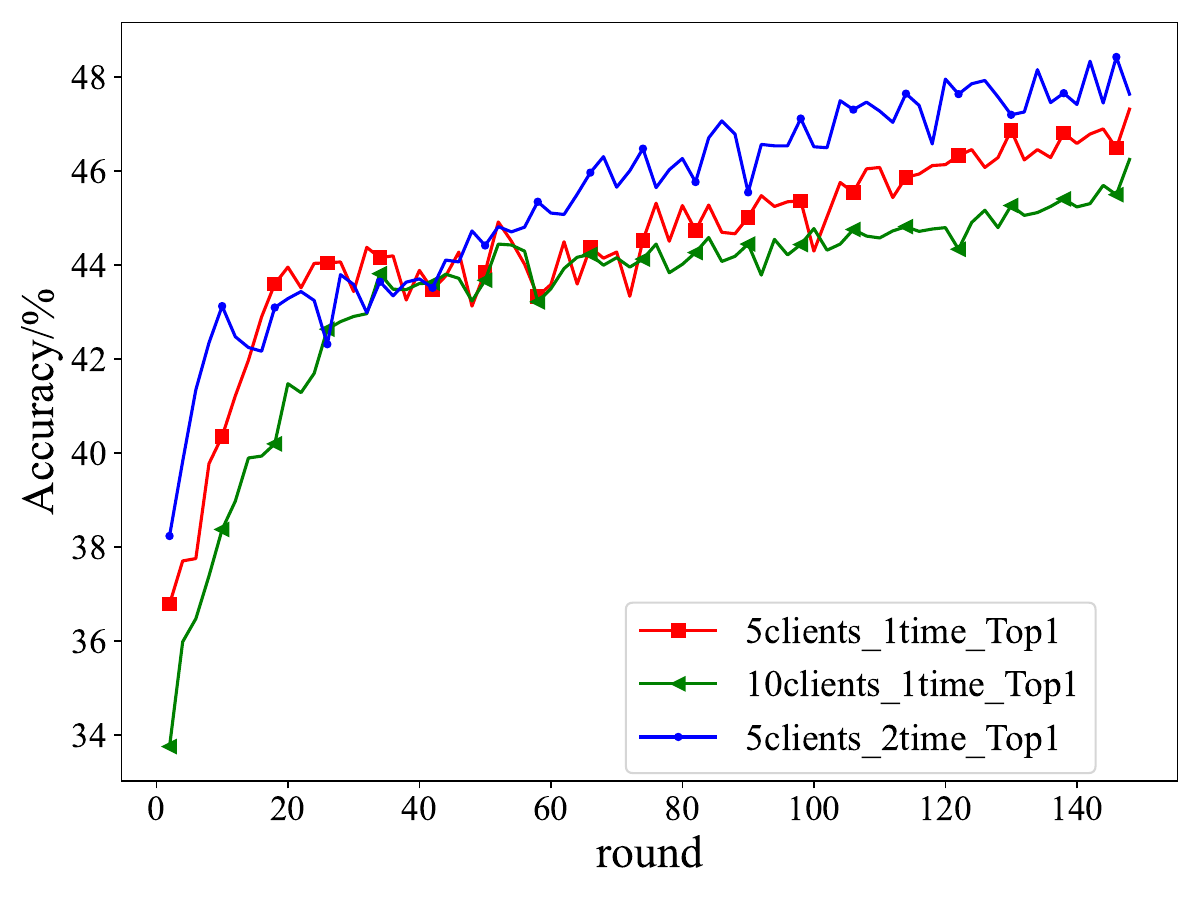}
		\label{fig9a}
	}\hfill
	\subfloat[Loss function]{%
		\includegraphics[width=0.48\linewidth]{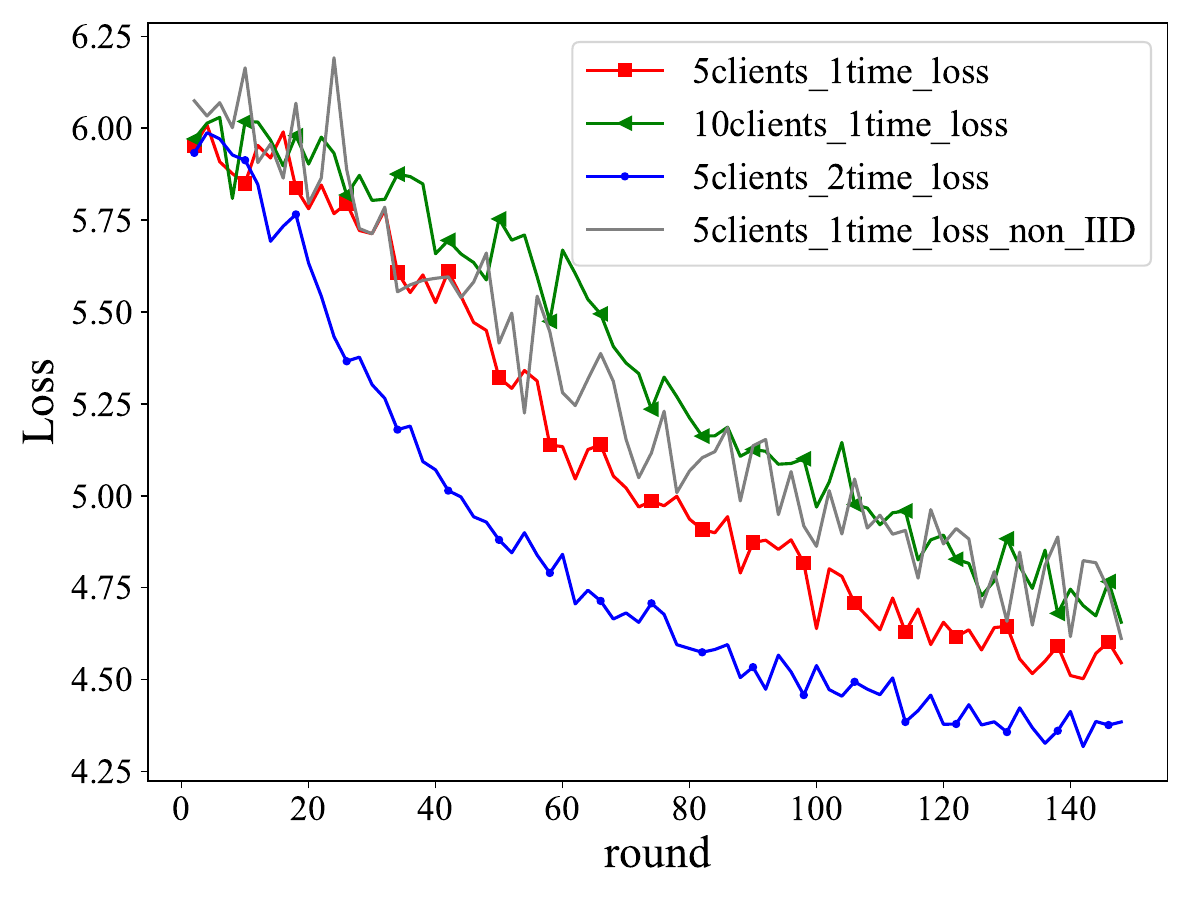}
		\label{fig9c}
	}
	\caption{Comparison of accuracy and loss}
	\label{fig9}
\end{figure*}

\textcolor{red}{In Fig. \ref{fig9}, we analyze scenarios where 5 and 10 vehicles participate in each training round. The red and green lines in Fig. \ref{fig9a} illustrate scenarios with 5 and 10 training vehicles, respectively. The red and blue lines in Fig. \ref{fig9a} correspond to scenarios where 5 vehicles participate in each round, with the red line representing a single local iteration and the blue line representing two local iterations. When 10 vehicles participate in each round of FLSimCo, the initial accuracy is the lowest. However, as training progresses, the accuracy gradually aligns with that achieved with 5 vehicles. This pattern is observed because, initially, the vehicles contribute a more diverse set of datasets. As iterations increase, the newly added vehicles bring increasingly similar datasets, reducing overall diversity. In this context, aggregating a smaller number of models in the initial rounds proves beneficial as it captures a broader range of data diversity. As shown in Fig. \ref{fig9c}, we compare the loss curves of the trained models from Fig. \ref{fig9a} with the loss function of 5 vehicles after one round of training on Non-IID datasets. Each experiment's loss function shows a downward trend, with the fastest convergence rate and lowest of loss achieved when 5 vehicles participate in each training round, performing 2 local iterations.}

\textcolor{red}{Notably, compared to the IID training set curves, the loss function on Non-IID datasets exhibits significant fluctuations in the early stages. As training continues, the loss function's trend becomes more similar to that of the IID datasets. This can be partly explained by the characteristics of Non-IID datasets, where the uneven distribution and diversity of data present greater challenges during training. As a result, the model's loss function fluctuates considerably at first but stabilizes gradually as it adapts to the data distribution over time. Additionally, conducting multiple rounds of local training allows the model to better learn data features, leading to lower loss values in a shorter period.}

\begin{figure}
	\center
	\includegraphics[scale=0.5]{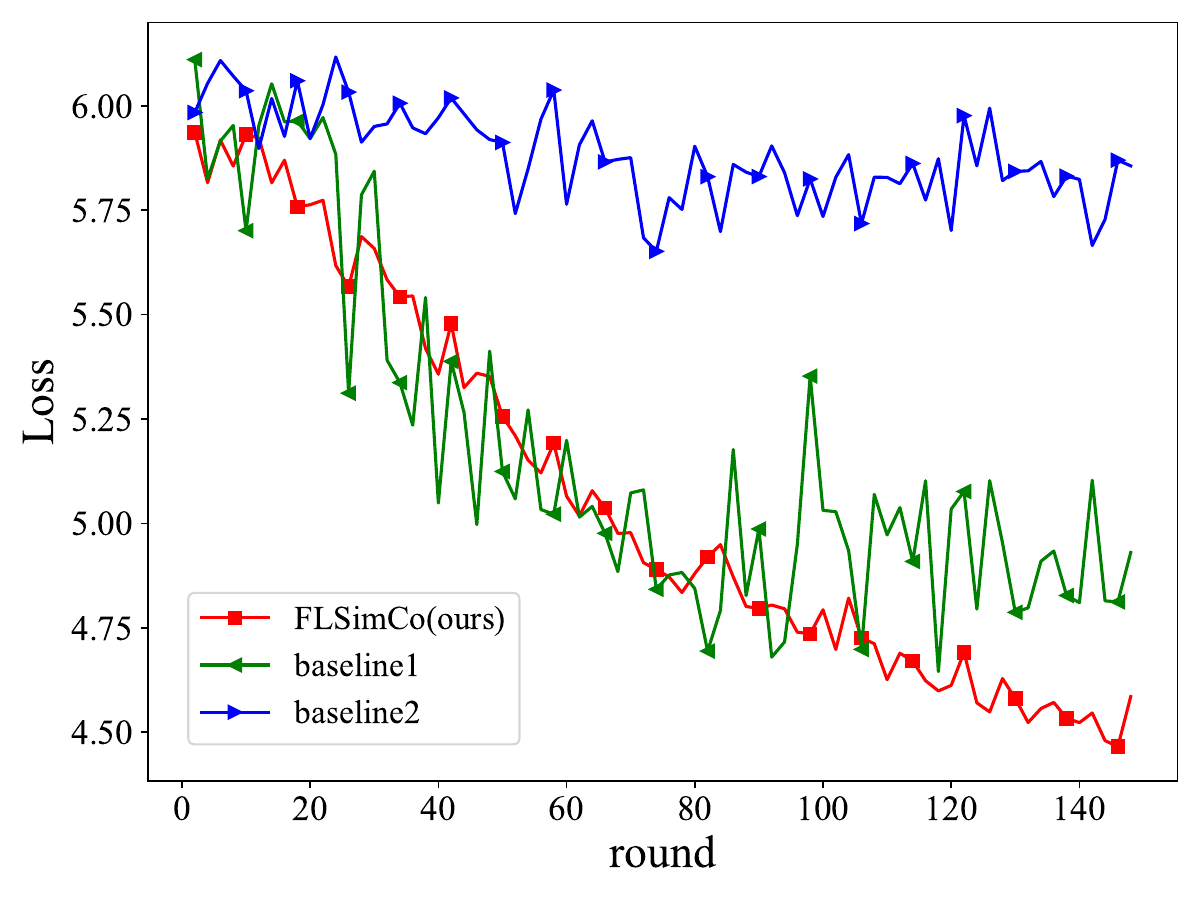}
	\caption{Aggregated with different weights}
	\label{fig8}
\end{figure}

In Fig. \ref{fig8}, we compare the loss function of different aggregated methods. We assume that some images will be blurred when the vehicle velocity exceeds $100 km/h$. To better demonstrate its performance, we also introduce two baseline algorithm. FedAvg is employed as baseline1, which averaging the model parameters accordingly \cite{hmcmahan2017}. Baseline2 indicates that RSU will discard the models trained by the vehicle velocity exceeding $100 km/h$, that is, discard the local model trained with the blurred images, and then use the FedAvg to aggregate model parameters.

\textcolor{red}{From the experimental results of baseline1, it can be observed that if local models trained on motion-blurred images are aggregated at RSU using indiscriminate FedAvg aggregation, these models negatively impact the global model, as evidenced by significant fluctuations in the loss curve. This occurs because models trained on motion-blurred images generally exhibit lower quality and poorer feature representation capabilities, leading to inaccurate or inconsistent gradient information. When these low-quality models are directly and uniformly integrated into the global model, their inaccurate gradient information destabilizes the learning process of the global model, resulting in pronounced fluctuations in the loss function. From the experimental results of baseline2, it can be seen that its loss curve converges the to slow and no significant downward trend. This is primarily because baseline2 discards some models during FedAvg aggregation, thus using fewer local models for aggregation. This approach, on the one hand, reduces the coverage of training images and, on the other hand, decreases the diversity and amount of information available for global model training. Consequently, the model fails to fully leverage the features from the local models of various vehicles during aggregation, which affects the global model's learning effectiveness. Therefore, even though local models trained on motion-blurred images may negatively impact the global model during aggregation, they still contribute valuable feature information. Proper handling and weight adjustment can enable these low-quality models to play a positive role in the global model aggregation process, thereby enhancing the performance and convergence speed of the global model. Hence, to mitigate the negative impact of these low-quality models on the global model, it is essential to adjust their weights in the model aggregation process.} Our proposed aggregation method assigns smaller weights to models trained by faster vehicles. It can be seen that the proposed algorithm can effectively reduce the fluctuation of the loss function, while facilitating the loss function and converge to a smaller value faster. This demonstrated that proposed approach can reduce the effect of image blur and also increase the global model training speed.\textcolor{red}{ From the standard deviation of the gradients of the three curves, it can be observed that our proposed FLSimCo method has a gradient standard deviation of 0.067, whereas baseline1 and baseline2 have gradient standard deviations of 0.23 and 0.10, respectively. Consequently, our method reduces the gradient standard deviation by 70.9\% and 33\%, respectively. These results indicate that, compared to existing baseline methods, our approach demonstrates a significant advantage in terms of gradient stability, effectively reducing the magnitude of gradient fluctuations during model training, and thereby facilitating a more stable and faster convergence process.}

\section{Conclusion}\label{sec6}

In this paper, we proposed a FLSimCo algorithm. Firstly, we addressed the dependency of supervised learning on labeled data by utilizing a DT-based SSL method, significantly reducing the cost of manual labeling. Additionally, we proposed SSL within the framework of FL that not only safeguarded vehicle privacy but also got generalized model. Lastly, considering that images captured by moving vehicles can suffer from motion blur, which negatively impacts the global model during aggregation, we incorporate the blur level as a weighting factor in the aggregation process. The contributions of this paper can be summarized as follows:

	\begin{itemize}
		
		\item \textcolor{red}{ In vehicle scenarios with relatively uniform driving environments, the DT-based self-supervised learning method demonstrates superior classification accuracy. Compared to the original method, it improves classification accuracy by 13.03\% on IID datasets and by 8.2\% on Non-IID datasets.}
		
		\item \textcolor{red}{In each round of FL, the fewer the participating vehicles, the lower the data diversity, resulting in higher initial classification accuracy. Additionally, increasing the number of local iterations can further improve classification accuracy. }
		
		\item \textcolor{red}{To address the potential negative impact of local models trained on blurred images during the FL aggregation process, assigning lower weights to models with higher blur levels can facilitate faster and more stable convergence of the loss function. Compared to the original method, the standard deviation of the loss function gradients is reduced by 70.9\% and 33\%, respectively.}

	\end{itemize}
	
\bibliography{sn-bibliography}
\end{document}